\documentclass[12pt]{article}       
\usepackage{graphicx}
\usepackage[hidelinks]{hyperref}

\usepackage{amsmath}
\usepackage{bm}
\usepackage{bbm}
\usepackage{amssymb,amsfonts}
\usepackage{subcaption}

\makeatletter
\newcommand{\removelatexerror}{\let\@latex@error\@gobble}
\makeatother
\usepackage[ruled,vlined,linesnumbered]{algorithm2e}

\usepackage[title]{appendix}

\DeclareMathOperator*{\argmin}{\arg\!\min}
\DeclareMathOperator*{\argmax}{\arg\!\max}

\begin{document}

\title{Constrained Clustering and Multiple Kernel Learning
without Pairwise Constraint Relaxation
}


\author{Benedikt Boecking\textsuperscript{1} \and Vincent Jeanselme\textsuperscript{1,2} \and Artur Dubrawski\textsuperscript{1}
\\
\\
\small
\textsuperscript{1}~Auton Lab, School of Computer Science, Carnegie Mellon University\\
\small
\textsuperscript{2}~MRC Biostatistics Unit, University of Cambridge
}



\date{}

\maketitle
\begin{abstract}
Clustering under pairwise constraints is an important knowledge discovery tool that enables the learning of appropriate kernels or distance metrics to improve clustering performance. These pairwise constraints, which come in the form of must-link and cannot-link pairs, arise naturally in many applications and are intuitive for users to provide. However, the common practice of relaxing discrete constraints to a continuous domain to ease optimization when learning kernels or metrics can harm generalization, as information which only encodes linkage is transformed to informing distances. We introduce a new constrained clustering algorithm that jointly clusters data and learns a kernel in accordance with the available pairwise constraints. To generalize well, our method is designed to maximize constraint satisfaction without relaxing pairwise constraints to a continuous domain where they inform distances. We show that the proposed method outperforms existing approaches on a large number of diverse publicly available datasets, and we discuss how our method can scale to handling large data.
\end{abstract}

\section{Introduction}
An important but sometimes neglected aspect of clustering is the impact of the underlying notion of similarity, e.g.\ an implicit assumption that the Euclidean distance is a good metric when applying the \textit{k}-means algorithm. 
Clustering is an inherently under-specified problem where the notion of a correct grouping depends on its context.  Thus, it is often unclear how to choose an appropriate similarity measure for a clustering task. While a user of a clustering algorithm may have an intuitive understanding of which instances should belong to the same clusters, it is generally difficult to map this intuition onto a metric or a feature set that would reflect such intuition well.

Clustering under pairwise constraints is an important knowledge discovery tool, and pairwise constraints in this setting enable learning of kernels or distance metrics to improve clustering performance. A number of algorithms have been developed that can simultaneously adapt the underlying notion of similarity or distance while clustering the data. 
The pairwise linkage information is usually available in the form of constraints--one set of \textit{must-link} pairs and one set of \textit{cannot-link} pairs of data instances--and the resulting problem is generally referred to as constrained clustering or semi-supervised clustering \cite{Basu2002seed, klein2002instance, wagstaff2001constrained}.
Since true clusters are unknown apriori in many practical scenarios, leveraging such feedback can be intuitive and convenient, for example in assessing  interpatient similarity~\cite{wang2011integrating}, or in information retrieval~\cite{cohn2003semi}. 
And, in many applications, pairwise constraints arise naturally as linkages between objects, providing an obvious way to encode meta-knowledge. In e.g.\ protein function prediction, the constraints can represent functional relationships between proteins~\cite{eisenberg2000protein}. In other applications, spatial or temporal proximity of samples can be used to induce such constraints, as in e.g.\ the analysis of spectral information from planetary observations~\cite{wagstaff2002intelligent}, in video segmentation and speaker identification~\cite{bar2003learning}, and face clustering in videos~\cite{wu2013constrained}. 

When pairwise linkage constraints are available, one not only desires a grouping of data that minimizes violations of known constraints, but a model that generalizes well to constraints beyond the observed ones. 
This is evident in existing work on constrained clustering with metric or kernel learning (e.g.~\cite{basu2006probabilistic,kulis2009semi}), 
where performance improvement is measured via predicted cluster membership on data for which constraints are not known upfront. 
To obtain objective functions that are easier to optimize in practice, formulations of constrained clustering with joint metric or kernel learning (e.g.~\cite{basu2006probabilistic,bilenko2004integrating,yin2010semi}) relax pairwise linkage constraints to a continuous space where they inform distances, even if the constraints only inform cluster membership and not relative distances between samples.  
The goal of learning a better metric or kernel is then formulated as making must-link pairs nearby and cannot-link pairs distant according to the resulting metric, which is a proxy to learning a clustering model that encourages the must-link pairs to belong to the same cluster and cannot-link pairs to different clusters.
Unfortunately, this relaxation can lead to over-specified constraints, e.g. a must-link constraint for samples that naturally lie on opposite ends of a cluster under a sensible metric. 



To uncover patterns in data that generalize well to unseen pairwise constraints, we introduce a simple but efficient constrained clustering algorithm that jointly learns a cluster model and kernel by maximizing the number of satisfied training constraints--without having to relax pairwise constraints to a continuous domain.  To this end, we use known pairwise constraint information to: (1) improve cluster initialization (2) learn a kernel by measuring constraint satisfaction when it is used for unconstrained clustering. The proposed method belongs to the category of soft-constrained clustering algorithms, which allow for the violation of some known constraints. 

The simple motivating idea is to use constraints to estimate how well a kernel uncovers the structure underlying known constraints when used for clustering.
We demonstrate how to learn a kernel from a set of bases by using a kernel \textit{k}-means algorithm and sparse Multiple Kernel Learning (MKL).
Our choice of kernel learning for the proposed method is deliberate. The kernel trick leads to the underlying clustering algorithm implicitly operating in some (possibly highly dimensional) feature space, allowing discovery of nonlinear cluster shapes. Furthermore, kernel methods can readily handle a variety of data types such as distributions \cite{poczos2012nonparametric},
time series \cite{cuturi2011fast}, trees \cite{croce2011structured}
, or graphs~\cite{Vishwanathan2006Fast} and application specific kernel families have been developed such as for object recognition \cite{sahbi2011context}. 
Additionally, MKL naturally allows the use of multiple views and transformations of the same data since kernels can be applied to varying views or feature sets. Prior research has shown the benefits of MKL over picking a kernel via cross-validation as well as benefits of sparse MKL formulations~\cite{gonen2011multiple,subrahmanya2010sparse}. 

We conduct experiments on 146 publicly available benchmark datasets~\cite{Olson2017} and demonstrate that our proposed approach performs better than popular alternatives on a large variety of data. 
We also show empirically that several existing approaches frequently converge to sub-optimal metrics, i.e.\ by using the proposed method one can find a better solution using the same type of distance metric and the same training data. 
Our results demonstrate that relaxing pairwise constraint labels to distance information in a continuous space can frequently yield sub-optimal pairwise metrics. 
We also show that the proposed algorithm can scale well to large datasets, which is not the case for many alternative methods. 
All of our code is open-sourced to ensure that our results can be readily reproduced
\footnote{Code available at \url{https://github.com/autonlab/constrained-clustering}}. 
We note that in this paper, we use `pairwise metric' as a generic term for distance, similarity, or dissimilarity function.

\section{Related Work}
\label{sec:relatedwork}

\textit{Semi-supervised clustering}--also called \textit{constrained clustering}--uses small amounts of weak label information which frequently comes in the form of \textit{pairwise constraints}~\cite{Basu2002seed, basu2004probabilistic, bilenko2004integrating, klein2002instance, wagstaff2000clustering, wagstaff2001constrained}. For clustering with pairwise constraints, supervision is provided in the form of \textit{must-link} (\textit{ML}) and \textit{cannot-link} (\textit{CL}) constraints which indicate same or different cluster membership of pairs of samples. Constrained clustering algorithms generally belong to  constraint-based and/or distance-based approaches, where the former do not include learning of an underlying pairwise metric. 
While out of scope of this article, constrained clustering has also been studied under the availability of cluster-level constraints~\cite{davidson2007complexity}, for scenarios where constraints are obtained from different sources \cite{bai2020semi}, and in settings where small sets of cluster label information are available~\cite{daume2006bayesian, finley2005supervised, liu2017partition}.

In constraint-based algorithms, a constraint sensitive assignment of samples to clusters is performed, in order to reduce violations of known constraints. 
Here, the literature often differentiates between hard and soft versions of constraint imposition, where the latter allow for some violations. 
In a constrained \textit{k}-means algorithm, the constraint sensitive assignment may lead to better cluster centers and therefore to better test set performance (as in e.g.~\cite{pelleg2007k}). 

Purely distance-based algorithms such as Mahalanobis Metric Learning for Clustering (MMC)~\cite{xing2003distance} and Information-Theoretic Metric Learning (ITML)~\cite{davis2007information} separate metric learning from the clustering step. MMC learns a Mahalanobis metric by minimizing the sum of squared distances between similar pairs under the constraint that the sum over dissimilar pairs is kept above some constant. ITML  learns a Mahalanobis distance metric that is close to a given initial one, and uses slack variables to keep distances between similar pairs within some margin while maintaining a greater margin between dissimilar pairs. While some authors refer to pairwise constraints as similar/dissimilar points, these pairs are generally assumed to stem from the same/different cluster. 

The exclusion of unlabeled data from the metric learning step motivated the introduction of joint metric learning and clustering via pairwise constraints. Seminal work are the Hidden Markov Random Field (HMRF) \textit{k}-means~\cite{basu2004probabilistic} and Metric Pairwise Constrained \textit{k}-means (MPC-Kmeans)~\cite{bilenko2004integrating} which  jointly learn a metric and cluster assignment via pairwise constraints. 
HMRF \textit{k}-means can be adapted to learn a variety of distortion measures including Bregman divergences. MPC-Kmeans, which is closely related to HMRF \textit{k}-means, learns a cluster-specific Mahalanobis distance which allows for clusters to lie in different subspaces. This concept of learning cluster-specific metrics was later also suggested for use in HMRF \textit{k}-means~\cite{basu2006probabilistic}. 

Researchers have  studied joint kernel learning and constrained clustering in both parametric and nonparametric ways. \cite{yan2006adaptive} derive an adaptive semi-supervised kernel \textit{k}-means algorithm (Adaptive-SS-Kernel-KMeans) inspired by HMRF \textit{k}-means. It learns kernel parameters such as the scale of the Gaussian kernel. In \cite{hoi2007learning}, the authors propose a nonparametric approach to learn a kernel matrix using pairwise constraints. The optimization problem is set to learn a kernel matrix that is consistent with known constraints while simultaneously being consistent with an assumed known similarity function. \cite{anand2013semi} introduce a semi-supervised kernel mean shift clustering (SKMS) algorithm. For the kernel learning step, SKMS updates an initial kernel matrix to meet specified target distance values to make pairs of samples with ML/CL constraints similar/dissimilar. Like ITML, SKMS uses slack variables during this step to relax the exact distance requirement.

Joint clustering and metric learning formulations based on or similar to HMRF \textit{k}-means iteratively adapt a metric or kernel according to a cluster loss as well as scaled penalties for violating the constraints (e.g.~\cite{basu2006probabilistic,yan2006adaptive}). Adapting the pairwise metric to reduce cluster loss may allow constraint information to be propagated to good initial cluster assignments. But it could also reinforce false cluster assignments. 
The inclusion of the cluster loss in pairwise metric learning also means that careful measures need to be taken to avoid trivial solutions. For example, \cite{yan2006adaptive} aim to avoid degenerate solutions by adding a constraint to the optimization problem such that the sum of distances of all samples to a random point is greater than some constant.
 
The use of pairwise constraints to learn improved embeddings for clustering of large datasets with deep neural networks has also been explored. 
\cite{hsu2016neural} design a loss function to train neural networks using pairwise constraints, and~\cite{hsu2017learning} devise a method to perform transfer learning on unknown classes and datasets using pairwise constraints and neural networks. 
\cite{fogel2019clustering} propose to learn an autoencoder and decoder using pairwise constraints to obtain an embedding for non-centroid based clustering by optimizing a representation loss, a reconstruction loss, and the pairwise loss. 
Naturally, all these approaches require large datasets to yield reliable models as well as a particular design of an appropriate network architecture that fits a given task. 

\section{Methodology}\label{sec:methodology}
We write vectors $\bm x$ in bold and matrices $\bm X$ in bold capital letters. We are given a dataset $\bm X \in \mathbb{R}^{n \times d}$ of $n$ samples and a small number of pairs of samples
$(\bm x_i,\bm x_j)$ known to be in either the same cluster identified by the set $\mathcal{M}$ of \textit{must-link} (ML) constraints , or in different clusters identified by the set $\mathcal{C}$ of \textit{cannot-link} (CL) constraints. These pairwise constraints may be weighted with $\omega_{ij}$ to reflect uncertainty about the relationship. Our goal is to guide the clustering in a way that minimizes constraint violations. Importantly, we want the cluster model to generalize to unseen constraints.  

\subsection{A Multiple Kernel Learning Algorithm}

We now introduce a constrained clustering algorithm we name KernelCSC which learns a linear combination kernel. 
The overall objective of this algorithm is to find a kernel which leads to an unconstrained clustering that maximizes satisfaction of known pairwise constraints. 
Pseudo code can be found in Algorithm~\ref{Algo1}. 
At each iteration of the algorithm, we obtain a candidate kernel $\bm K$ parameterized by a vector $\bm \beta$. Using $\bm K$, we cluster the data via kernel \textit{k}-means, and evaluate the resulting clusters $\hat{S}$ as a function of satisfied pairwise constraints such that the reward function is:
\begin{equation}\label{eq:objective}
\resizebox{.9\textwidth}{!}{%
  $\begin{aligned}
  R(\hat{S})=&\frac{1}{|\mathcal{M}|+|\mathcal{C}|}\left(\sum_{(\bm x_i,\bm x_j)\in \mathcal{M}}\omega_{ij}\mathbbm{1}[l_i  = l_j ] + \sum_{(\bm x_i,\bm x_j)\in \mathcal{C}}\omega_{ij}\mathbbm{1}[l_i \neq l_j ]\right).
  \end{aligned}$%
  }
\end{equation}
where $\hat{S}$ is is the proposed clustering, and $l_i$ is the cluster label assigned to sample $i$ in $\hat{S}$. As common in the related semi-supervised clustering literature, we multiply a pairwise constraint by a weight $\omega_{ij}$ if provided. 

\textbf{Linear combination kernel:} 
We use the Multiple Kernel Learning (MKL) paradigm to define the kernel matrix $\bm K$.
For now, assume that we have means to obtain a candidate parameter vector $\bm\beta$ defining our kernel during each iteration of Algorithm~\ref{Algo1}. Let $ \mathcal{G} =  \{\bm G_i\}_{i=1}^p, \bm G_i \in \mathbb{R}^{n \times n},\bm G_i\succeq 0 \  \forall i \in \{1,\dots,p\}$ be a set of $p$ Kernel matrices. Given $\bm \beta \in \mathbb{R}_{+}^p$, we can create a linear combination kernel $\bm K = \sum_{i=1}^p \beta_i \bm G_i$. Note that we constrain our analysis and experiments to linear combination kernels, but that nonlinear combination kernels can easily be plugged into the algorithm as well.  

\textbf{Kernel k-means:} 
Once $\bm K$ is defined, the unconstrained \textit{kernel k-means} clustering step partitions data into $k$ disjoint sets $\mathcal{S} = \{S_1, S_2,\dots, S_k\}$ for the simplified objective:
\begin{equation}\label{eq:kernelkmeanssimple}
    \argmin_{\{S_c\}_{c=1}^k}\  tr(\bm K) - \sum_{c=1}^k \frac{\sum_{\*x_i,\*x_j\in S_{c}} \bm K_{ij}}{|S_c|},
\end{equation}
where $S_c$ is a set containing all elements assigned to cluster $c$. To make good use of known constraints, we initialize centroids using the \textit{farthest first} scheme~\cite{kulis2009semi}. This serves to provide better initial cluster assignments  and more stable clusters for similar kernels across iterations. 

\textbf{Optimizing $\bm \beta$:} Whether Algorithm~\ref{Algo1} performs well depends on an effective acquisition function, which is responsible for identifying promising values of $\bm\beta$. We can view steps 5-8 of Algorithm~\ref{Algo1} as a function $f$ of $\bm\beta$. That is, $f(\bm\beta)$ constructs the kernel $\bm K$, performs kernel k-means, and returns the reward of the resulting partition via Eq.~\ref{eq:objective}. Since this involves a clustering step, $f$ cannot be differentiated and is also somewhat expensive to evaluate. Thus, we require an effect gradient-free optimization procedure to find good candidates for $\bm\beta$. One important empirical observation is that the best $\bm\beta$ is highly likely to be sparse since base kernels are constructed via heuristics which are not guaranteed to lead to reasonable clustering results themselves. Thus, dense $\bm\beta$ generally lead to bad groupings. We therefore constrain the search space to sparse candidates $\mathcal{D}=\{\bm \beta : \bm \beta \in [0,1]^p, ||\bm \beta||_0\leq c \}$. 

A naive way of optimizing $\bm \beta$ is to sample uniformly over $\mathcal{D}$ in each iteration of the algorithm. The more sophisticated approach is to use Sequential Model Based Optimization (SMBO) principles such as introduced in~\cite{hutter2011sequential}. At each iteration, we fit a model $g$ to the history $\mathcal{H}=\{(\bm\beta_1,f(\bm\beta_1)),\dots,(\bm\beta_{t-1},f(\bm\beta_{t-1})) \}$ of previously explored $\bm\beta$ values. The model $g$ represents our prior belief about our true function $f$ over the domain and is used as an approximation to find promising candidate values. We optimize the Upper Confidence Bound~\cite{srinivas2010gaussian} over the domain of sparse vectors to obtain the next candidate:
\begin{equation}
    a(\bm\beta| \mathcal{H}) = \argmax_{\bm \beta \in \mathcal{D}} \mu(\bm\beta|\mathcal{H})+\kappa * \sigma(\bm\beta|\mathcal{H}) 
\end{equation}
where we obtain the posterior mean $\mu(\bm\beta|\mathcal{H})$ and standard deviation $\sigma(\bm\beta |\mathcal{H})$ from $g$ and optimize by drawing values uniformly over $\mathcal{D}$ and taking the best sample. That is, we fit a regressor $g$ to $\mathcal{H}$ and then obtain $\mu$ and $\sigma$ from the regressor for samples in $\mathcal{D}$ to find the sample maximizing the acquisition function. SMBO is well suited to applications such as ours, in which optimizing $g$ is less computationally expensive than optimizing $f$ directly. Random Forest (RF) regression or Gaussian Process are common choices for $g$.
\begin{figure}[!t]
 \removelatexerror
  \begin{algorithm}[H]
    \SetAlgoLined
    \SetKwInOut{Input}{Input}\SetKwInOut{Output}{Output}
    \Input{$\mathcal{G}=\{\bm G_1,\dots,\bm G_p\}$: base kernel matrices; $k$: number of clusters;  $\mathcal{M},\mathcal{C}$: must-link and cannot link constraint sets; $\mathcal{W}$: weights of pairwise constraints; $a$: acquisition function}
    \Output{$\hat{S}^{best},\bm \beta^{best},y^{best}$}
    $\bm C_M$= \textit{ConnectedComponents}($\mathcal{M}$),\;
    $\mathcal{H}=\varnothing$\; %
     \While{stopping criterion not met}{
        $\bm\beta \xleftarrow{} \argmax_{\bm \beta \in \mathcal{D}} a(\beta|\mathcal{H}) $ \;
        $\bm K \xleftarrow{} \sum_{i=1}^p \bm \beta_i \bm G_i$\;
        $S_{init} \xleftarrow{} $ \textit{FarthestFirst}($\bm C_M,\mathcal{M},\mathcal{C},\bm K$)\;
        $\hat{S} \xleftarrow{}$ \textit{KernelKMeans}($\bm K,S_{init},k$)\;
        $y \xleftarrow{} R(\hat{S})$\;
        $\mathcal{H} = \mathcal{H} \cup (\bm\beta,y) $\;
        update  $y^{best}$, $\bm \beta^{best}$, $\hat{S}^{best}$\;
     }
     \caption{\textbf{Constraint Satisfaction Clustering}}\label{Algo1}
    \end{algorithm}
\end{figure}

\subsubsection{Complexity and Scalability}

Assuming a negligible cost to the gradient free optimization which generates $\bm \beta$ candidates, the proposed method requires $\mathcal{O}(pn^2)$ storage and  $\mathcal{O}(n^2(dp+t))$ computation for creating kernels and running the algorithm, with $n$ samples in $d$ dimensions, $p$ base kernels, and $t$ optimization iterations. We can scale the algorithm to larger datasets by approximating the feature map for kernel functions, e.g. by using the Nystr\"{o}m method~\cite{gittens2016revisiting} leading to $\mathcal{O}(qn)$ memory and $\mathcal{O}(n(q^2p+t))$ computation where $q$ is the rank of the approximation. It has been shown that the use of Nystr\"{o}m approximations for kernel \textit{k}-means is theoretically sound, practically useful, and scalable to large data sets~\cite{wang2019scalable}.  We provide results showing the consistency of such an approximation and feasible runtime on a large dataset in Section \ref{sec:experiments}. In addition, for large datasets  one may choose to downsample data for which no constraints are known to further reduce training complexity while learning an appropriate kernel.

\subsubsection{Optional Constrained Clustering Step with Fixed Kernel}

Once Algorithm \ref{Algo1} has terminated, a user may choose to perform a final constrained clustering with a fixed kernel. While optional, this can be done to satisfy more training constraints if such behavior is desirable in a particular application. We define the objective function of this  \textit{constrained kernel k-means} by adding the following penalty term $g(\bm K,\{S_c\}_{c=1}^k)$ to Eq.~\ref{eq:kernelkmeanssimple}:
\begin{equation}\label{eq:ckkmeanspenalty}
    \resizebox{.7\textwidth}{!}{%
  $\begin{aligned}
  g(\bm K,\{S_c\}_{c=1}^k) = \sum_{(\*x_i,\*x_j)\in \mathcal{C}}\omega_{ij}\mathbbm{1}[l_i = l_j ]\left(\bm K_{ii}-2 \bm K_{ij}+\bm K_{jj} \right)\\ + \sum_{(\*x_i,\*x_j)\in \mathcal{M}}\omega_{ij}\mathbbm{1}[l_i \neq l_j ]\left(D_{max}- \bm K_{ii}+2\bm K_{ij}-\bm K_{jj} \right),
  \end{aligned}
  $%
  }
\end{equation}
where $D_{max} = \max \left(\{\bm K_{ii}-2\bm K_{ij}+\bm K_{jj} \} _{i,j=1}^n \right)$ is the largest distance in the feature space. 
Pairwise constraint violation costs are scaled by distances in feature space to obtain penalties that are of similar magnitude to the distances we observe between samples and cluster centers. 
This allows outliers to violate constraints. Centers are again initialized using the \textit{farthest-first} algorithm and clusters are learned via an iterative EM-like algorithm with a greedy approach to handle constraint dependencies as in~\cite{basu2004probabilistic}. Note that, as in related work such as HMRF \textit{k}-means~\cite{basu2004probabilistic} and MPC-Kmeans~\cite{bilenko2004integrating}, this soft constrained formulation of kernel \textit{k}-means does not guarantee the satisfaction of all training constraints. 
\subsubsection{Implementation Details}
To create the base set of kernels for our proposed algorithm we use Radial Basis Function (RBF), Laplace, Polynomial, Sigmoid, and Linear kernels. We compute each kernel on the raw data as well as on standardized data. For the parameters of each kernel, we adopt standard heuristics to create grids of reasonable values. For the width parameter of the RBF kernel, we estimate the median of all pairwise Euclidean distances and multiply this value by different scaling factors. Similarly, for the Laplacian kernel we use multiples of the inverse of the approximate median Manahattan distance. For Sigmoid and Polynomial kernels we use the approximate median of the inner product. We compute polynomial kernels of degree $2$ and $3$. We scale each kernel matrix in our set of kernels by a positive scalar to avoid numerical issues. To optimize $\bm \beta$, we use a RF regressor as our model $g$ and set $\kappa=1.0$ as values in a wide range around it worked well across variety of datasets. For a fair comparison without fine-tuning, we set the sparsity parameter $c$ to fixed low value of 5 for all datasets. 

\section{Experiments}\label{sec:experiments}

\begin{figure}[t]
  \includegraphics[height=150px]{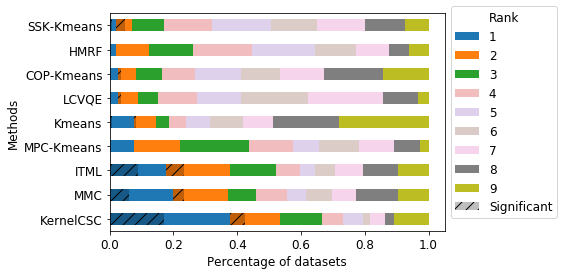}
  \centering
  \caption{Ranks of algorithms on all 146 datasets, based on mean ARI.  Shading indicates significant difference at $\alpha=0.05$  normal confidence intervals. Ties are resolved by assigning the minimum rank.
  }
  \label{fig:rankAdjRand}
\end{figure}
\subsection{Datasets and Algorithms}
We use the Penn Machine Learning Benchmarks database (PMLB)~\cite{Olson2017} to comprehensively evaluate the proposed approach on a large number of publicly available benchmark problems covering a wide range of applications\footnote{Data: \url{https://github.com/EpistasisLab/penn-ml-benchmarks}}. We limit the analysis presented in this paper to all labeled datasets in PMLB containing more than $99$ samples. This results in a collection of 146 datasets. We show comparisons to algorithms considering scalability and the representativeness of the approach in regard to the frameworks available in the vast semi-supervised clustering literature. Further, to allow for a fair comparison, we constrain the analysis to algorithms where the number of clusters $k$ is assumed to be known. 
In addition to \textbf{\textit{k}-means}, we run the following algorithms on the standardized data:

\noindent \textbf{COP-Kmeans}~\cite{wagstaff2001constrained}: a hard-constrained \textit{k}-means algorithm aiming to resolve all constraint violations. 

\noindent \textbf{LCVQE}~\cite{pelleg2007k}: a soft constrained \textit{k}-means which does not terminate if constraints are violated. 

\noindent  \textbf{SSK-Kmeans}~\cite{kulis2009semi}: a constrained graph clustering algorithm; we use cross-validation to choose an RBF kernel to create the input affinity matrix. 

\noindent \textbf{HMRF \textit{k}-means}~\cite{basu2006probabilistic} and \textbf{MPC-Kmeans}~\cite{bilenko2004integrating}:  semi-supervised clustering algorithms that also perform joint metric learning.

\noindent \textbf{ITML}~\cite{davis2007information} and \textbf{MMC}~\cite{xing2003distance}: metric learning algorithms. For both, we use LCVQE to partition the data with the learned metric.\footnote{We evaluated a range of alternatives to establish hard baselines. A final LCVQE partitioning provided the best performance compared to  other options such as \textit{k}-means or COP-Kmeans.} 

hich provided better performance than a subsequent k-means clustering (or its variants such as COP Kmeans)

\subsection{Experimental Setup and Results}

Our experiments follow conventions established in related work. Algorithms are applied to the full data, but training constraints are only available between samples in a small train set, while performance is only measured on samples belonging to a test set. Training constraints are sampled uniformly at random from the binary adjacency matrix of points belonging to the training set.
All algorithms are trained and evaluated on the exact same sets of constraints and test points. 
We compared all algorithms across a range of evaluation metrics including Normalized Mutual Information, Adjusted Mutual Information, Adjusted Rand Index, Fowlkes–Mallows Index, and F-score. Due to space constraints, we illustrate test set performance using Adjusted Rand Index (ARI) scores only, but note that the relative performance differences and overall conclusions were consistent when we used other evaluation metrics as noted. We provide additional Figures using other evaluation metrics in Appendix~\ref{appendixA}.

\subsubsection{Fixed Number of Training Constraints}\label{sec:fixedresults}

We repeat the following procedure $10$ times for each dataset: we create a stratified random split of the data based on the true cluster label, designating $25\%$ as a training set. We randomly select $10\%$ of all possible pairs (up to a maximum number of $5000$ pairs) in the training data to obtain known constraints. We augment constraints by transitive and entailed constraints. Wherever algorithms consider weights for constraints, we assign unit weights. For our KernelCSC method, we set the maximum number of optimization iterations to $1000$.

\begin{figure*}[ht]
    \centering
  \includegraphics[width=\textwidth]{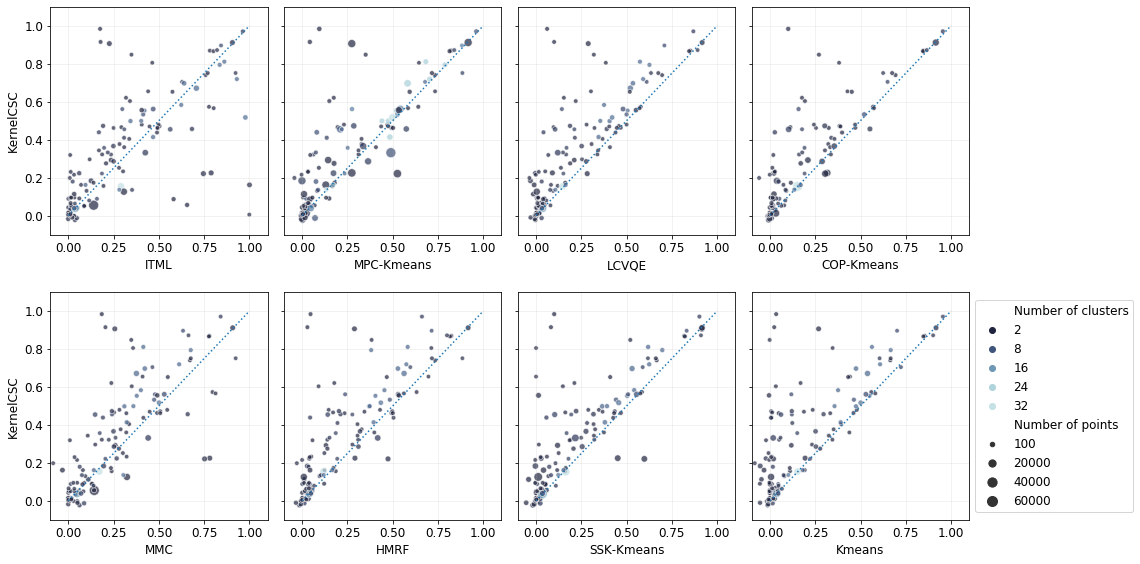}
  \centering
  \caption{Mean test-set performance (ARI) of the proposed method (y-axes) against considered alternatives (x-axes), over all $146$ datasets. When our method performs better, points lie above the diagonal. Size and shading of each point indicate the diversity of dataset characteristics. }
  \label{fig:pairAdjRand}
\end{figure*}

The scatter plots in Fig.~\ref{fig:pairAdjRand} summarize the mean test set ARI across all datasets, showing that our proposed method outperforms other algorithms on a large number of datasets. In Fig.~\ref{fig:rankAdjRand}, we display a summary of the ranks that each algorithm achieves on all datasets on the basis of the mean ARI over random trials. The proposed algorithm places first for $37.7\%$ of the datasets, and at least second in $53.4\%$. For the top $3$ ranks, this figure also displays the percentage of datasets for which the difference in mean ARI to all lower ranked algorithms is significant, calculated via normal confidence intervals over the random runs at $\alpha=0.05$. MMC achieves the second most top ranks, placing first in $17.1\%$ of the datasets, and at least second in $21.9\%$, closely followed by ITML which obtains more significant top results than MMC and more second places. We note that--as far as we have tested--increasing the size of the training data and/or increasing the number of known constraints in the training fold, did not affect the results in a way that would change the overall conclusions. When the number of optimization iterations of our algorithm is set to $1000$, we find that random parameter optimization works as well as the proposed SMBO strategy. 
\begin{figure}[t]
  \includegraphics[height=150px]{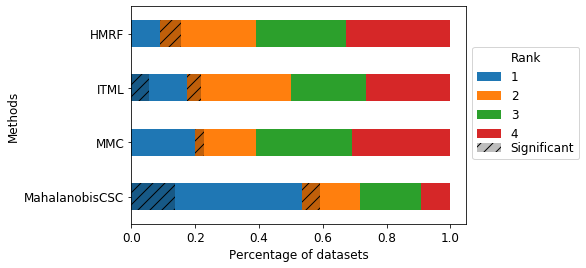}
  \centering
  \caption{We also adapt our algorithm (CSC) to learn a diagonal Mahalanobis metric instead of a linear combination kernel. The bars summarize ranks achieved over all 146 datasets using mean ARI with methods using Mahalanobis metric. The results indicate that the performance improvements also observed in the MKL version stem from better generalization of learned pairwise metrics by measuring constraint satisfaction.
  }
  \label{fig:bayesianMahalanobis}
  
\end{figure}

An alternative to the MKL based version of our algorithm proposed in this paper is to learn a Mahalanobis distance in conjunction with \textit{k}-means. There are several disadvantages to an approach based on learning a Mahalanobis metric leading the kernel version being our preferred approach, the main concern being that gradient free optimization becomes increasingly difficult for high dimensional datasets. However, to show that the improved clustering performance does not stem from non-linearities introduced by using kernels or the particular bases we chose but rather from better generalization of the learned similarity function, we adapt our approach to also learn a Mahalanobis distance. Instead of a linear combination kernel, we here learn a diagonal projection matrix to transform the data and perform clustering via k-means instead of kernel k-means. The vector which parameterizes this Mahalanobis distance is again learned via SMBO, but without the sparsity restriction we use for multiple kernel learning. Fig.~\ref{fig:bayesianMahalanobis} provides a relative comparison to methods from the literature that also learn a Mahalanobis metric, showing that our approach outperforms them on a large number of datasets when learning a Mahalanobis metric. Since we are using the same training data and metric, Fig.~\ref{fig:bayesianMahalanobis} indicates that related methods frequently converge to sub-optimal Mahalanobis metrics which do not generalize as well to unseen constraints. In our experiments, our proposed methods of learning a kernel (KernelCSC) outperformed the alternative of learning a Mahalanobis metric (MahalanobisCSC) on $63.7\%$ of the datasets. 

We also compare KernelCSC which uses MKL to the approach of choosing one kernel from our set of base kernels via cross-validation. The objective here remains the same (Equation \ref{eq:objective}), but kernel learning is replaced with simply picking one base kernel. 
The results we obtained showed that MKL using a simple linear combination kernel learning approach can indeed boost performance, giving a higher mean ARI on $69.2\%$ of datasets. 
\subsubsection{Increasing Size of Training Set}\label{sec:evolutionresults}
\begin{figure}[t]
  \includegraphics[height=150px]{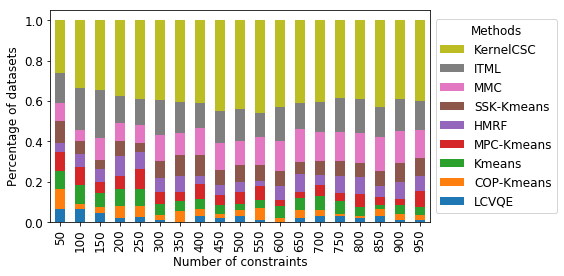}
  \centering
  \caption{Percentage of times over all datasets each algorithm is ranked first on the test set (y-axis), vs.\ the number of pairwise training constraints (x-axis) used in training. The ranks were established on test-sets using mean ARI over 10 random trials.
  }\label{fig:evolutionFirstAdjRand}
\end{figure}
To study how relative test set performance evolves as more training constraints become available, we  randomly draw training constraints from a train partition in a range from $50$ to $1000$ pairs, in increments of $50$. Again, we repeat these experiments $10$ times where $75\%$ of each dataset is held out for testing and training constraints are augmented by transitive and entailed constraints. Due to the large number of experiments conducted, we set the maximum number of optimization iterations of our KernelCSC method to $100$.
In Fig.~\ref{fig:evolutionFirstAdjRand}, we summarize the performance of all algorithms by the percentage of datasets where each places first, showing that our method outperforms all others according to this evaluation, regardless of the number of known constraints. The experiments also reveal that our method achieves good relative performance even with a small number of optimization iterations.  
We also find that, once the metric is learned, the optional  soft constraint clustering step  does not significantly impact measured performance on the test sets compared to using the learned kernel with an unconstrained kernel k-means to obtain the final partition.

\subsubsection{Scalability to Large Datasets}

\begin{figure}[t]
  \includegraphics[height=150px]{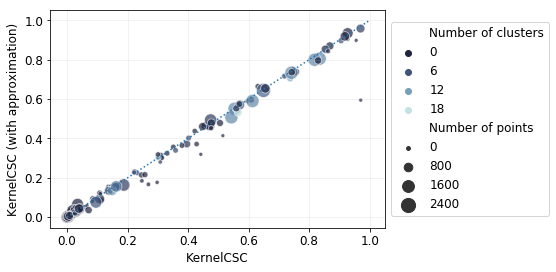}
  \centering
  \caption{A scatter plot comparing the ARI performance of KernelCSC (x-axis) to its scalable implementation using kernel approximations (y-axis), based on test set performance.
  }
  \label{fig:scale}
  
\end{figure}
Here we show that our KernelCSC method can scale well to large datasets and that the required approximations do not decrease performance considerably. Fig.~\ref{fig:scale} shows that an implementation of KernelCSC using Nystr\"{o}m approximations to each kernels' feature map produces results very similar to the exact implementation. 
One of the datasets contained in the PMLB benchmark database is the large \textit{kddcup} dataset which contains $494020$ points with $23$ clusters. We are able to cluster this dataset using the approximate KernelCSC without any multi-processing (Intel Xeon Gold 6152 CPU), with $1000$ optimization iterations and $5000$ known constraints in $\sim 396$ minutes. We used $62$ base kernels and each of the kernels was approximated with $150$ components.   

\section{Discussion}\label{sec:discussion}

Our experiments show that the proposed method typically prevails over popular alternatives on a wide variety of data. 
An experiment into the evolution of mean test ARI per number of known training constraints shows the superior performance even if a small number of constraints are known, after only a few kernel learning iterations. Our approach relies on MKL to learn a kernel and can find good solutions despite relying on a gradient-free optimization. Further, the small number of base kernels which are created automatically for each dataset based on common heuristics appear to work well out of the box for a large variety of datasets. 

The proposed approach can learn a kernel without relaxing pairwise constraints, and our experiments suggest that the frequently superior performance is the result of optimizing for a pairwise metric that--when used for clustering--is expected to generalize well to unseen pairwise constraints. In one set of experiments (Fig.~\ref{fig:bayesianMahalanobis}), we optimize a diagonal Mahalanobis metric in a similar vein to our proposed method by swapping the kernel matrix and kernel \textit{k}-means with a simple \textit{k}-means used on a learned projection of the data. This approach is not expected to work well out of the box for data with a large number of features since the function domain becomes more difficult to optimize with gradient-free algorithms. 
Yet, the approach outperforms related methods that also learn Mahalanobis metrics, on a large number of datasets. This experiment highlights that related approaches with relaxed constraints  frequently converge to sub-optimal metrics when constraints are obtained from the true underlying clusters. 

We believe that there are several core issues that lead to some methods being outperformed by our proposed approach. First, many related methods are formulated to adapt a pairwise metric to decrease the distance between ML pairs and to increase distance between CL pairs. This relaxation of linkage information to distance information needs to be considered carefully. It is important to observe that the pairwise linkage constraints that guide learning generally do not encode how similar or dissimilar the pairs are but merely inform cluster membership. It is possible that algorithms overfit to pairwise constraint information when they are relaxed to a continuous space, even when slack variables are used.
Second, objective functions in constrained clustering with joint metric learning often combine a clustering loss-- e.g.\ cluster variance~\cite{basu2004probabilistic,bilenko2004integrating,kulis2009semi,yan2006adaptive}--and constraint violation cost of relaxed pairwise constraints. One issue is that the cluster variance minimization is performed indiscriminately for all data points in a cluster including ones that violate constraints, which may reinforce sub-optimal solutions especially during early iterations. Further, this aspect of the objective can be decreased by simply shrinking the distances between all points, which necessitates additional steps to avoid trivial solutions. Finally, due to the cluster assignment of a sample being a function of the cluster loss and penalty terms that depend on known constraints, the cluster assignment of a point with known constraints can be different from the assignment of an equivalent point without known constraints. 

Our proposed Sequential Model Based Optimization strategy finds good solutions quickly, but we also find that random parameter optimization over the sparse constrained space provides a good alternative. The random parameter optimization generally finds good solutions within hundreds of iterations. This strategy is especially useful for smaller datasets where the evaluation of the objective function is cheap, while SMBO is well suited to very large datasets where fewer premeditated function evaluations are desired.

\section{Conclusion}\label{sec:conclusion}

We introduced a new algorithm for constrained clustering with kernel learning. It uses discrete pairwise membership constraints to guide learning. We conducted experiments on $146$ datasets that demonstrated superior performance of the proposed approach compared to popular alternatives,  highlighting the importance of generalization to unseen constraints in designing constrained clustering algorithms.
When pairwise constraints only indicate same or different cluster membership, a relaxation to an encoding of distances--while convenient for optimization--can easily lead to constraints being over-specified. 

Advantages of our proposed method are that--due to the use of MKL--it is straightforward to incorporate data from multiple views, and that it naturally extends to problem settings where data is not available in a simple tabular form such as in time series, or distributions. In addition, the proposed method scales gracefully to handling large datasets. 

The proposed algorithm has several limitations in its present form. Keeping a number of Gram matrices or kernel approximations in memory may in practice require substantial amount of these resources. Further, our reliance on gradient-free optimization limits the number of bases in MKL or dimensionality of the data for Mahalanobis metric learning that can be handled comfortably. 
Promising adaptations of our approach include learning nonlinear combination kernels, exploring alternatives to the currently used gradient free optimization, and evaluating utility of our approach in semi-supervised multi-view settings.
\section*{Acknowledgements}
This work was partially supported by a Space Technology Research Institutes grant from NASA’s Space Technology Research Grants Program and by Defense Advanced Research Projects Agency’s award FA8750-17-2-0130.

\newpage
\bibliographystyle{spmpsci}      
\bibliography{mybib.bib}   

\newpage
\begin{appendices}
\section{Varying Evaluation Metrics}\label{appendixA}
\begin{figure*}[ht]
    \centering
  \includegraphics[height=150px]{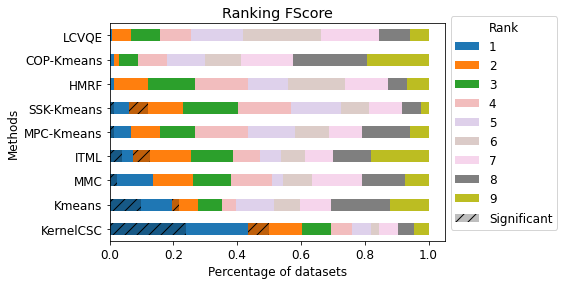}
  \centering
  \caption{Percentage of times over all datasets each algorithm is ranked first on the test set (y-axis), vs.\ the number of pairwise training constraints (x-axis) used in training. The ranks were established on test-sets using mean F-score over 10 random trials.
  }
  \label{fig:evolutionFirstFscore}
\end{figure*}

\begin{figure*}[ht]
    \centering
  \includegraphics[height=150px]{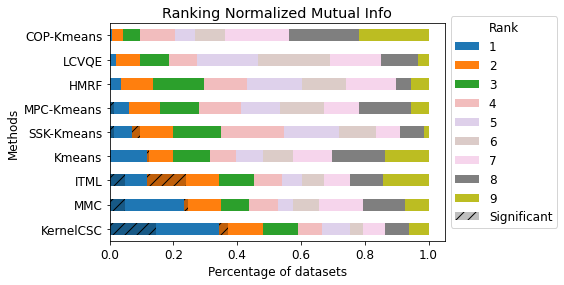}
  \centering
  \caption{Percentage of times over all datasets each algorithm is ranked first on the test set (y-axis), vs.\ the number of pairwise training constraints (x-axis) used in training. The ranks were established on test-sets using mean Normalized Mutual Information over 10 random trials.
  }
  \label{fig:evolutionFirstNMI}
\end{figure*}

To evaluate and compare the performance of the proposed approach we computed a number of metrics established in related literature such as the Adjusted Rand Index (ARI), F-score and Normalized Mutual Information (NMI). As stated previously, the superior performance of our approach remains consistent when we evaluate relative performance according to other evaluation metrics. Figure~\ref{fig:evolutionFirstFscore} and Figure~\ref{fig:evolutionFirstNMI} demonstrate that the proposed approach also outperforms related approaches under F-score and NMI. Under NMI, the percentage of datasets where the proposed approach ranks first is slightly lower compared to evaluations done with F-score and ARI. 

\section{Informativeness and Coherence Measures}\label{appendixB}
\cite{davidson2006measuring} suggested that averaging over different randomly chosen constraint sets may mask interesting properties of the individual constraint sets. The authors introduce two quantitative measures, \textit{informativeness} and \textit{coherence}, and among other things, use these measures to inspect disparities in performance of different clustering algorithms. Since our evaluation averages algorithm performance over randomly chosen constraint sets, we here discuss an additional analysis we performed, inspecting the relative performance when using more and less informative or coherent constraint sets, to see if differences in the relative ranking of algorithms emerge when these metrics vary.

\begin{figure}[ht]
\begin{subfigure}[t]{.5\textwidth}
  \centering
  \includegraphics[height=100px]{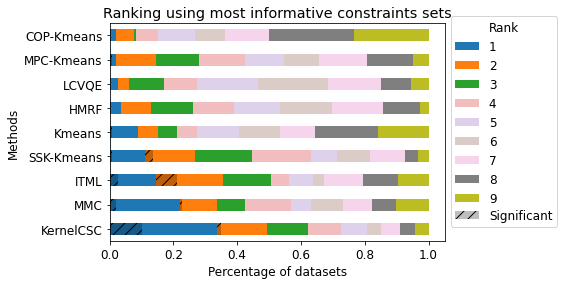}  
  \caption{Mean over the most informative trials.}
\end{subfigure}
\begin{subfigure}[t]{.5\textwidth}
  \centering
  \includegraphics[height=100px]{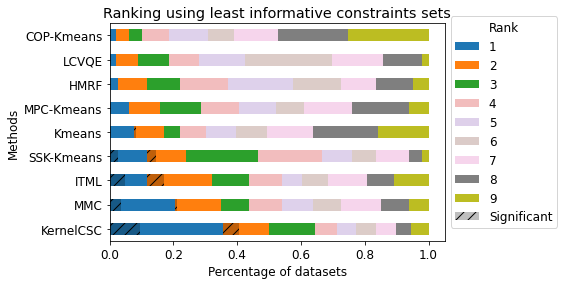} 
  \caption{Mean over the least informative trials.}
\end{subfigure}
\caption{Percentage of times over all datasets each algorithm is ranked first on the test set (y-axis), vs.\ the number of pairwise training constraints (x-axis) used in training. The ranks were established on test-sets using mean Adjusted Rand Index over the 5 out of 10 most and least informative random trials.}
\label{fig:evolutionFirstInfo}
\end{figure}

\begin{figure}[ht]
\begin{subfigure}{.5\textwidth}
  \centering
  \includegraphics[height=100px]{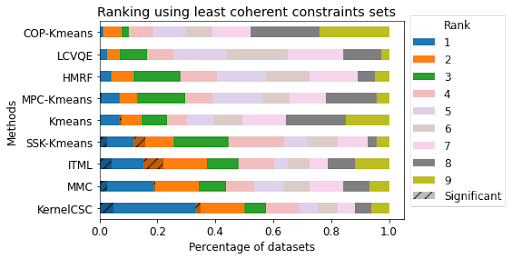}   
  \caption{Mean over the most coherent trials.}
\end{subfigure}
\begin{subfigure}{.5\textwidth}
  \centering
  \includegraphics[height=100px]{results/CoherentMin.png} 
  \caption{Mean over the least coherent trials.}
\end{subfigure}
\caption{Percentage of times over all datasets each algorithm is ranked first on the test set (y-axis), vs.\ the number of pairwise training constraints (x-axis) used in training. The ranks were established on test-sets using mean Adjusted Rand Index over the 5 out of 10 most and least coherent random trials.}
\label{fig:evolutionFirstCoherence}
\end{figure}

It turns out that when we compare relative algorithm performance across datasets using different levels of  informative and coherent constraint sets, we observe no systematic differences in the relative performance of the top performing algorithms. We do see some drop in performance across all algorithms when using less coherent constraints. In the relative comparison between algorithms, this does lead to some ranking differences being less significant. To illustrate this, we recreate Figure~\ref{fig:rankAdjRand} using the 5 least informative constraint sets compared to the 5 most informative constraint sets in Figure~\ref{fig:evolutionFirstInfo}, and similarly showing the 5 least coherent constraint sets compared to the 5 most coherent constraint sets in Figure~\ref{fig:evolutionFirstCoherence}. All rankings in these figures rely on Adjusted Rand Index scores.

\end{appendices}

\end{document}